\documentclass[lettersize,journal]{IEEEtran}
\usepackage{cite}
\usepackage{amsmath,amssymb,amsfonts}
\usepackage{algorithmic}
\usepackage{graphicx}
\usepackage{textcomp}
\usepackage[hidelinks,colorlinks=false]{hyperref}
\usepackage{algorithm}
\usepackage{booktabs}
\usepackage{multirow}

\usepackage{scalerel}
\usepackage{tikz}
\usetikzlibrary{svg.path}

\definecolor{orcidlogocol}{HTML}{A6CE39}
\tikzset{
  orcidlogo/.pic={
    \fill[orcidlogocol] svg{M256,128c0,70.7-57.3,128-128,128C57.3,256,0,198.7,0,128C0,57.3,57.3,0,128,0C198.7,0,256,57.3,256,128z};
    \fill[white] svg{M86.3,186.2H70.9V79.1h15.4v48.4V186.2z}
                 svg{M108.9,79.1h41.6c39.6,0,57,28.3,57,53.6c0,27.5-21.5,53.6-56.8,53.6h-41.8V79.1z M124.3,172.4h24.5c34.9,0,42.9-26.5,42.9-39.7c0-21.5-13.7-39.7-43.7-39.7h-23.7V172.4z}
                 svg{M88.7,56.8c0,5.5-4.5,10.1-10.1,10.1c-5.6,0-10.1-4.6-10.1-10.1c0-5.6,4.5-10.1,10.1-10.1C84.2,46.7,88.7,51.3,88.7,56.8z};
  }
}

\newcommand\orcidicon[1]{\href{https://orcid.org/#1}{\mbox{\scalerel*{
\begin{tikzpicture}[yscale=-1,transform shape]
\pic{orcidlogo};
\end{tikzpicture}
}{|}}}}

\newcommand{\cmark}{\checkmark}
\newcommand{\xmark}{--}

\def\BibTeX{{\rm B\kern-.05em{\sc i\kern-.025em b}\kern-.08em
    T\kern-.1667em\lower.7ex\hbox{E}\kern-.125emX}}

\begin{document}
\title{LD-SLRO: Latent Diffusion Structured Light for 3-D Reconstruction of Highly Reflective Objects}
\author{Sanghoon Jeon$^{\textsuperscript{\orcidicon{0009-0006-4947-2051}}}$, Gihyun Jung$^{\textsuperscript{\orcidicon{0009-0009-4980-4708}}}$, Suhyeon Ka$^{\textsuperscript{\orcidicon{0009-0003-6629-3520}}}$, and Jae-Sang Hyun$^{\textsuperscript{\orcidicon{0000-0003-1711-8243}}}$, \IEEEmembership{Member, IEEE}
\thanks{This work has been submitted to the IEEE for possible publication. Copyright may be transferred without notice, after which this version may no longer be accessible. 
This research was supported by the Culture, Sports and Tourism R\&D Program through the Korea Creative Content Agency grant funded by the Ministry of Culture, Sports and Tourism in 2024 (Project Name: Global Talent for Generative AI Copyright Infringement and Copyright Theft, Project Number: RS-2024-00398413, Contribution Rate: 30\%), the National Research Foundation of Korea(NRF) grant funded by the Korea government(MSIT) (Project Number: RS-2025-16072782, Contribution Rate: 30\%), the Technology Innovation Program (Project Name: Development of AI autonomous continuous production system technology for gas turbine blade maintenance and regeneration for power generation, Project Number: RS-2025-25447257, Contribution Rate: 40\%) funded By the Ministry of Trade, Industry and Resources(MOTIR, Korea). Corresponding author: Jae-Sang Hyun. Sanghoon Jeon, Gihyun Jung, Suhyeon Ka, and Jae-Sang Hyun are with the Department of Mechanical Engineering, Yonsei University, Seoul 03722, South Korea. (e-mail: hyun.jaesang@yonsei.ac.kr)}
}

\maketitle

\begin{abstract}
Fringe projection profilometry–based 3-D reconstruction of objects with high reflectivity and low surface roughness remains a significant challenge. When measuring such glossy surfaces, specular reflection and indirect illumination often lead to severe distortion or loss of the projected fringe patterns. 
To address these issues, we propose a latent diffusion-based structured light for reflective objects (LD-SLRO). Phase-shifted fringe images captured from highly reflective surfaces are first encoded to extract latent representations that capture surface reflectance characteristics. These latent features are then used as conditional inputs to a latent diffusion model, which probabilistically suppresses reflection-induced artifacts and recover lost fringe information, yielding high-quality fringe images.
The proposed components, including the specular reflection encoder, time-variant channel affine layer, and attention modules, further improve fringe restoration quality. In addition, LD-SLRO provides high flexibility in configuring the input and output fringe sets.
Experimental results demonstrate that the proposed method improves both fringe quality and 3-D reconstruction accuracy over state-of-the-art methods, reducing the average root-mean-squared error from 1.8176\,mm to 0.9619\,mm.
\end{abstract}


\section{Introduction}
\label{sec:introduction}
\IEEEPARstart{3}{-D} Reconstruction methods are in high demand in industrial fields such as the automotive industry\cite{lin2018recognition}, automated manufacturing\cite{qian2021high}, and aerospace engineering\cite{von2016multiresolution}. Among these methods, fringe projection profilometry (FPP)\cite{xu2020status}, as a structured light method, is widely used due to its high speed, high resolution, high accuracy, simple setup, and non-contact operation\cite{zhang2018high-review}.
A typical FPP system consists of a projector and a camera: the projector illuminates the object with phase-shifted fringe patterns, and the camera captures the deformed patterns on the surface. The recorded fringe sequence is then used to estimate the phase and reconstruct the 3-D shape.

However, measuring highly reflective surfaces with FPP remains difficult. On such surfaces, the captured intensity is a mixture of diffuse reflection and strong specular components. Specular reflection distorts the fringe waveform and breaks the sinusoidal assumption of phase-shifting algorithms, while specular highlights frequently saturate the sensor and cause overexposed regions where fringe information is lost. These issues are most severe for mirror-like objects with low surface roughness. In addition, secondary reflections between nearby surface patches introduce interreflections, which further wash out or warp the projected fringes.
As a result, reliable 3-D measurement of mirror-like and electroplated surfaces remains challenging in practical industrial applications.

To address mirror-like surfaces, recent work in computer graphics has investigated multi-view reconstruction under physically based rendering equations\cite{verbin2024ref-nerf}. For instance, Liu et al.\cite{liu2023nero} reconstructed mirror-like objects by combining multi-view images with a microfacet BRDF formulation. While effective, such approaches typically require multiple cameras and careful multi-view calibration. They also rely on iterative optimization rather than a pre-trained data-driven model, which leads to high computational cost and long inference time, making them impractical for industrial use.

In FPP-based 3-D measurement, most existing work targets metallic parts that are reflective but sufficiently rough so that fringes remain observable. Traditional approaches to handle reflection-induced errors can be categorized into hardware-based and algorithm-based methods\cite{feng2018high-review}. Hardware-based methods modify the acquisition process, for example by using multiple exposures\cite{zhang2009high-multiexp} or adjusting the projected intensity to avoid saturation\cite{waddington2014modified}. Cao et al.\cite{cao2022high-transparent} proposed using a transparent screen as an optical mask to reduce highlights. Although these strategies can improve measurements, they often increase acquisition time, require additional hardware, or complicate system tuning. Algorithm-based methods instead apply post-processing without changing system parameters, such as using inverted fringe patterns\cite{jiang2016high-inverted} or exploiting color-channel separation\cite{yin2017high-color}. However, many of these methods require extra pattern captures, which can introduce motion-induced errors\cite{jeon2024motion}, or become sensitive to surface color.

Recently, deep learning has also been explored for accurate measurement in FPP\cite{zuo2022deep}. Zhang et al.\cite{zhang2020high-dl} showed that saturation under diffuse reflection can be reduced by increasing the number of phase steps, and proposed a convolutional neural network (CNN) that synthesizes additional phase-shifted fringes from a small input set. While this helps with diffuse saturation, it does not explicitly address distortions caused by specular reflection, which often dominate on mirror-like surfaces.
To address specular reflections, Yang et al.\cite{yang2022high-hdrnet} introduced a two-stage network for single-image fringe enhancement on reflective surfaces, where saturated and dark regions are first identified and then enhanced. It mainly targets exposure-related degradation and does not explicitly account for non-sinusoidal distortions that may persist even in well-exposed regions.
Li et al.\cite{li2022three-dcunet} combine DC-UNet-based highlight region extraction with a dilated residual network for pixel-wise correction to suppress highlights in fringe images before phase computation.
Song and Wang\cite{song2024ffc} propose Y-FFC, a Y-shaped network that leverages frequency-domain fast Fourier convolution and gradient-domain cues to suppress specular reflection induced non-sinusoidal distortions in fringe patterns. The method requires separating the fringe image into specular and diffuse reflection components for training and primarily exploits gradient information from the diffuse component. However, for mirror-like plated objects as shown in Fig.~\ref{setup}(b), the observed intensity is typically dominated by specular reflection, making reliable separation of the diffuse component difficult in practice.

To overcome above challenges, this article proposes a latent diffusion-based structured light for reflective objects (LD-SLRO), which simultaneously address overexposure, fringe contrast loss, and interreflection-induced distortions commonly observed on mirror-like surfaces.
Latent diffusion\cite{rombach2022high-ldm} is a generative model that learns a high-dimensional data distribution by iteratively denoising in a compact latent space.
We adopt this idea and condition the denoiser on reflection-related cues, which provides a practical way to capture the complex, multi-factor degradations of reflective-surface fringes (e.g., overexposure, contrast loss, and interreflection).
LD-SLRO operates in the latent space and consists of two encoders, one decoder, and an attention-based denoiser. In particular, the specular reflection encoder extracts reflection characteristics of the target surface, and a time-variant channel-affine fusion mechanism injects these cues into the denoiser over diffusion timesteps to jointly model specular effects and fringe structures for accurate 3-D measurement.
In addition, the framework decouples the input and output fringe configurations, enabling fast acquisition from a sparse input set while generating a denser output set with a different frequency configuration for high-quality supervision and measurement. Experiments show that LD-SLRO improves both fringe restoration quality and 3-D reconstruction accuracy over state-of-the-art methods, reducing the average root-mean-squared error from 1.8176\,mm to 0.9619\,mm.
The proposed multi-condition latent-diffusion mechanism separately encodes complementary observations and fuses them over diffusion timesteps, which can be adapted to other restoration problems involving heterogeneous inputs.

\section{Principle}

\subsection{Fringe Projection Profilometry}{

\begin{figure}[!t]
\centerline{\includegraphics[width=\columnwidth]{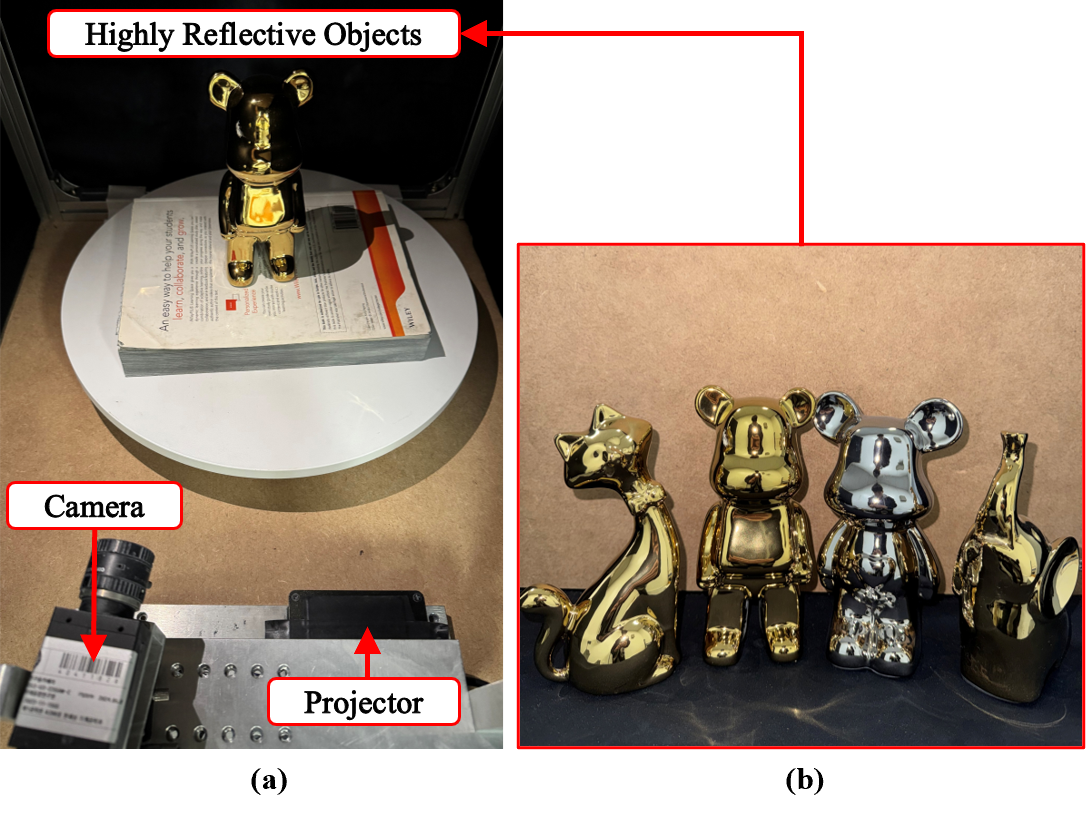}}
\caption{Experimental setup and test objects. (a) Experimental system setup of FPP. (b) Photograph of highly reflective objects.}
\label{setup}
\end{figure}

FPP typically consists of a single camera and a projector as shown in Fig.~\ref{setup}(a). The projector sequentially projects phase-shifted fringe patterns onto the object, and the camera captures the resulting deformed fringe images. The $i$-th captured fringe image of $N$ images can be represented as
\begin{equation}\label{eq:fpp_fringe}
I_i(x,y)=A(x,y)+B(x,y)\cos\bigl[\varphi(x,y)+\delta_i\bigr],\,i=1,2,\dots,N
\end{equation}
where $A$ is the average intensity, $B$ is the modulation amplitude, $\varphi$ is the phase, $\delta_i$ denotes the phase shift for the $i$-th image, and $N$ is the number of phase shift steps.
The phase is driven by a phase-shifting algorithm \cite{zuo2018phase} as
\begin{equation}
\varphi(x,y) = 
\arctan\!\left[
\frac{
\sum_{k=1}^{N} I_{k}(x,y)\,\sin\!\left(\frac{2\pi (k-1)}{N}\right)
}{
\sum_{k=1}^{N} I_{k}(x,y)\,\cos\!\left(\frac{2\pi (k-1)}{N}\right)
}
\right].
\label{eq:w_phase}
\end{equation}
Increasing the number of captured images $N$ improves measurement accuracy but reduces acquisition speed.
Since the phase is defined on the interval $(-\pi,\pi]$ and is spatially discontinuous, phase-unwrapping is required to obtain the absolute phase map in which the phase value is uniquely determined along the direction of phase variation, as follows:
\begin{equation}
    \Phi(x,y)=\varphi(x,y)+2\pi\times k(x,y)
\end{equation}
where $k$ is an integer, commonly known as the fringe order.
Temporal phase-unwrapping methods\cite{zhang2018absolute}, such as binary gray-code phase unwrapping\cite{cheng2024adaptive-graycode}, are widely used due to their robustness on high-contrast surfaces.
The absolute phase associated with each camera pixel enables camera–projector calibration\cite{zhang2006novel}, and three-dimensional depth is then recovered by triangulation using the calibrated geometric relationship among the camera, projector, and object.
}

\subsection{Specular Reflection Induced Error}{
\begin{figure}[!t]
\centerline{\includegraphics[width=\columnwidth]{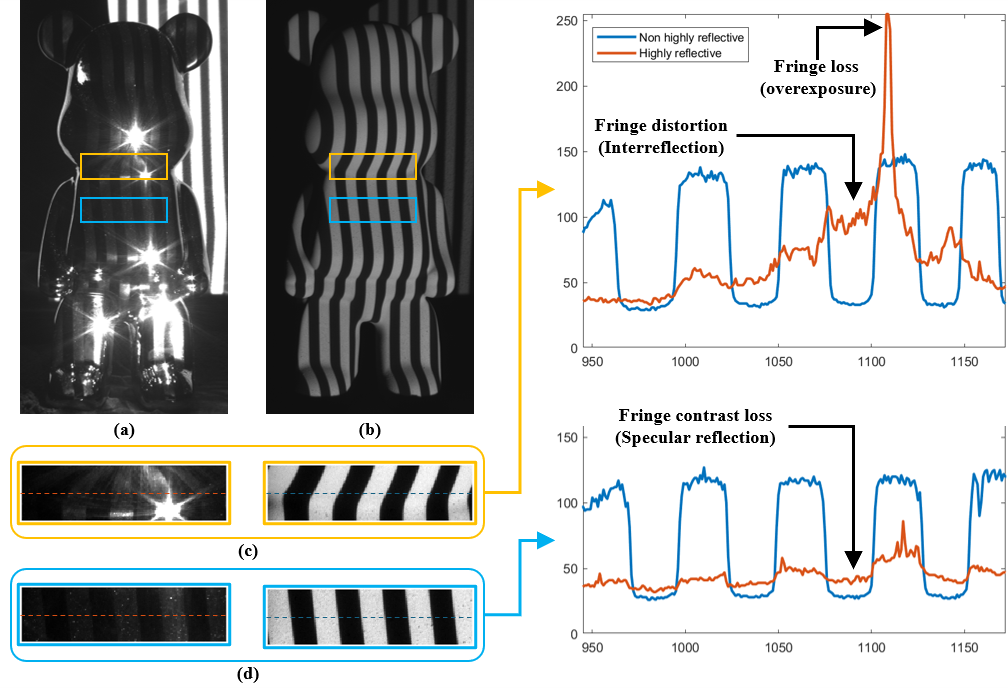}}
\caption{Typical fringe degradations on highly reflective surfaces. (a) Highly reflective fringe image (b) Non highly reflective fringe image (c) Fringe loss and distortion (d) Fringe contrast loss}
\label{fig:fringe_error}
\end{figure}
On highly reflective surfaces, specular reflection and interreflection introduce a strong global illumination component. The camera no longer observes only the direct structured pattern, but instead captures a superposition of the direct fringes and secondary reflections from neighboring surface patches. Since FPP relies on precise intensity modulation of the projected fringes, this additional component distorts the effective waveform and introduces phase errors, which in turn cause noticeable inaccuracies in the reconstructed 3-D shape.

In Fig.~\ref{fig:fringe_error}(a), the projected fringe pattern on a highly reflective plated surface is still visually recognizable, whereas Fig.~\ref{fig:fringe_error}(b) shows the corresponding pattern on a non–highly reflective object. However, their photometric behavior is markedly different. In highly reflective regions, strong specular highlights cause local overexposure. 
As shown in Fig.~\ref{fig:fringe_error}(c), the overexposed pixels appear as sharp spike-like intensity peaks along the line profile, indicating fringe loss, where the fringe waveform is locally clipped and the phase information is irrecoverably destroyed.

Even in non-overexposure regions, light reflected from one specular patch can illuminate neighboring patches and be reflected again before reaching the camera. This interreflection effect produces a spatially varying mixture of fringe signals with different phases, especially in concave geometries. Consequently, the observed fringe profile becomes locally warped and asymmetric, which appears as fringe distortion in Fig.~\ref{fig:fringe_error}(c), and cannot be described by simple additive noise.

Specular reflection also reduces the effective fringe modulation. Let $I_{\mathrm{d}}$ denote the diffuse component and $I_{\mathrm{s}}$ the non- or weakly-modulated specular component.
Locally, the captured intensity can be approximated as
\begin{equation}
    I(x,y) = I_{\mathrm{s}}(x,y) + I_{\mathrm{d}}(x,y)\bigl[1 + \gamma_0 \cos\varphi(x,y)\bigr],
\end{equation}
where $\gamma_0$ is the intrinsic fringe modulation on an ideal diffuse surface. The effective fringe modulation observed by the camera, defined in terms of the Michelson contrast $\gamma_{\mathrm{eff}}$, then becomes
\begin{equation}
    \gamma_{\mathrm{eff}} = \gamma_0 \frac{I_{\mathrm{d}}}{I_{\mathrm{d}} + I_{\mathrm{s}}},
\end{equation}
which decreases as $I_{\mathrm{s}}$ increases on the reflective surface. This reduction is visualized in Fig.~\ref{fig:fringe_error}(d) as a significant loss of fringe modulation compared to the diffuse case.

These observations motivate our multi-condition formulation and the dual-encoder conditioning, which separately captures fringe cues and reflection-related artifacts.
}

\section{Proposed Method}

\begin{figure*}[!t]
\centerline{\includegraphics[width=0.75\textwidth]{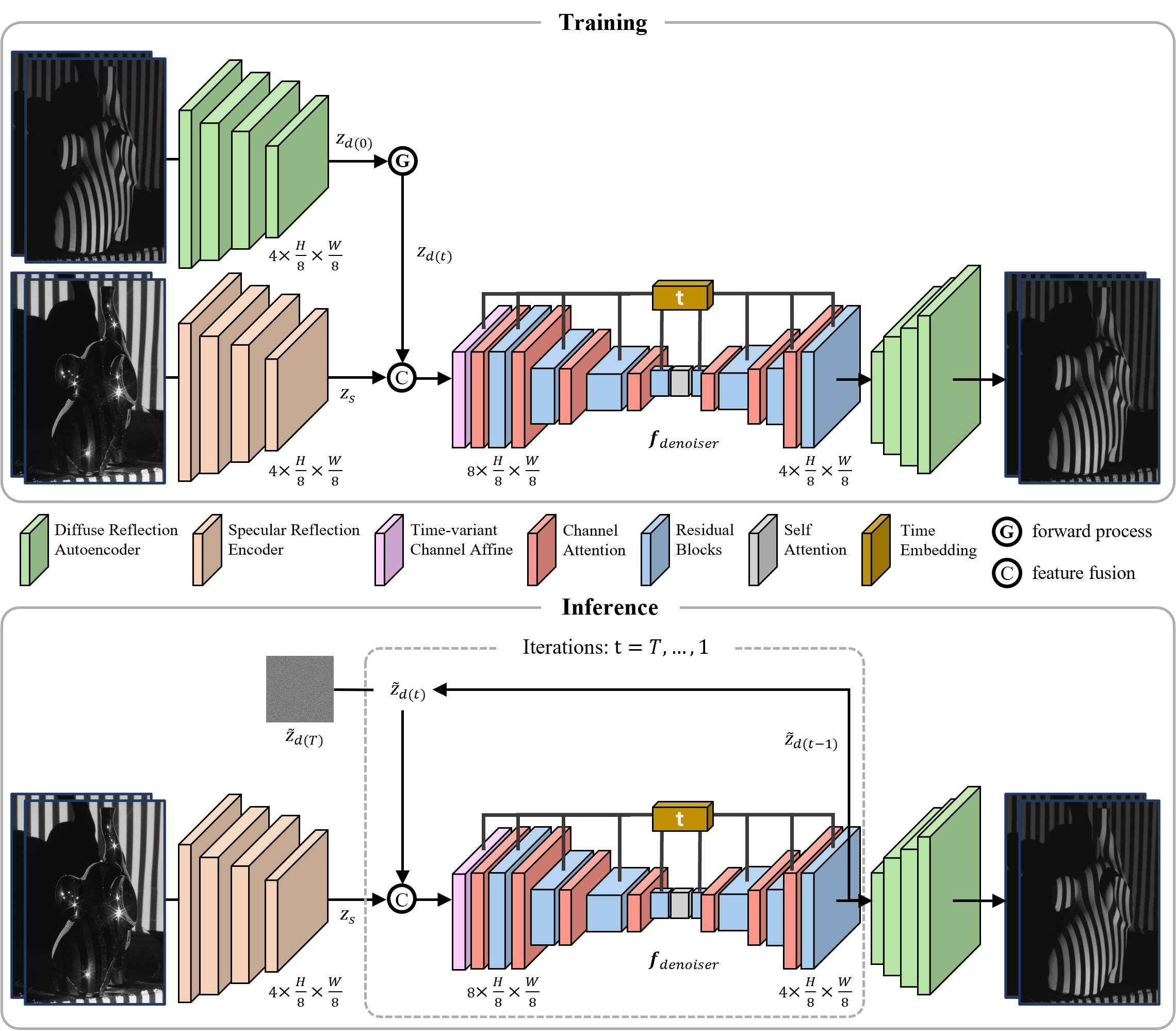}}
\caption{The network architecture of the proposed LD-SLRO consists of three parts: a denoiser network, a pretrained diffuse reflection autoencoder, and a pretrained specular reflection encoder.}
\label{network}
\end{figure*}

\subsection{Denoising Diffusion Probabilistic Model (DDPM)}{
DDPMs\cite{yang2023diffusion} are a class of generative models to model a high-dimensional data distribution by learning to reverse a predefined Markov forward diffusion process which gradually corrupts data by adding Gaussian noise. DDPM consists of a fixed forward noising process $q$ that progressively maps data $x_0$ to nearly isotropic Gaussian noise $x_T$, and a trainable reverse process $p_\theta$ that learns to denoise step-by-step from $x_T$ back to $x_0$. The forward process is a Markov chain:
\begin{align}\label{eq:foward_process}
q(x_{1:T}\mid x_0) &= \prod_{t=1}^T q(x_t\mid x_{t-1}),\\
q(x_t\mid x_{t-1}) &= \mathcal{N}(x_t;\sqrt{1-\beta_t}x_{t-1},\beta_t \mathbf{I})
\end{align}
where $\beta_t\in(0,1)$ is a noise variance schedule and $x_1,\ldots,x_T$ denote the intermediate states at time steps $t=1,\ldots,T$.
Since the forward process is Gaussian process, it is convenient to use the closed form that directly samples $x_t$ from $x_0$:
\begin{equation}\label{eq:simple_xt}
x_t = \sqrt{\bar{\alpha}_t}\,x_0 + \sqrt{1-\bar{\alpha}_t}\epsilon
\end{equation}
where $\alpha_t = 1 - \beta_t,\
\bar{\alpha}_t = \prod_{i=1}^t \alpha_i,\
\epsilon \sim \mathcal{N}(0,\mathbf{I})$.

The denoising Markov reverse process is modeled as a Gaussian conditional with learnable parameters:
\begin{align}\label{eq:reverse_process}
    p_\theta(x_{0:T}) &= p(x_T)\prod_{t=1}^T p_\theta(x_{t-1}\mid x_t),\\
    p_\theta(x_{t-1}\mid x_t) &= \mathcal{N}\!\big(x_{t-1};\,\boldsymbol{\mu}_\theta(x_t,t),\,\mathbf{\Sigma}_\theta(t)\big).
\end{align}
The training objective is to make $p_\theta(x_{t-1}\mid x_t)$ approximate the true posterior $q(x_{t-1}\mid x_t,x_0)$. Ho et al.\cite{ho2020denoising} let the neural network predict the noise $\epsilon_\theta (x_t,t)$ used to corrupt $x_0$ by minimizing the simplified training objective:
\begin{equation}\label{eq:SimpleLoss}
\mathcal{L}_{\mathrm{simple}}
= \mathbb{E}_{x_0\sim q(x_0),\;\epsilon\sim\mathcal{N}(0,\mathbf{I}),\;t\sim\mathrm{U}[1,T]}\big\lVert
\epsilon - \epsilon_\theta(x_t,t)\big\rVert^2.
\end{equation}

Alternatively, under the same parameterization the network can be trained to predict the clean sample $\hat{x}_0$ directly instead of the corruption noise. This is equivalent to predicting the noise which can be written as
\begin{equation}
    \hat{\epsilon}=\frac{x_t-\sqrt{\bar{\alpha}_t}\,\hat{x}_0}{\sqrt{1-\bar{\alpha}_t}}.
\end{equation}
In this article, we parameterize our network $\mathrm{x}_\theta(x_t,c,t)$, which is conditioned on the conditioning embedding $c$ and the timestep $t$, to predict $x_0$ directly by minimizing
\begin{equation}\label{eq:x0Loss}
\mathcal{L}_{\mathrm{x_0}}
= \mathbb{E}_{x_0\sim q(x_0),\;\epsilon\sim\mathcal{N}(0,\mathbf{I}),\;t\sim\mathrm{U}[1,T]}\big\lVert
x_0 - \mathrm{x}_\theta(x_t,c,t)\big\rVert^2
\end{equation}
because, in our latent-diffusion setup, the latent representation is relatively high-dimensional and the model has sufficient capacity to encode the clean sample. In such regimes the network can effectively store and recover $x_0$ from $x_t$, making direct reconstruction simpler and empirically more stable than learning a complex conditional noise distribution. 

Using the trained denoiser network, samples are generated by initializing \(x_T\sim\mathcal{N}(0,\mathbf{I})\) and iteratively applying the reverse update\cite{songdenoising} for \(t=T,\ldots,1\):
\begin{equation}
x_{t-1}
= \sqrt{\bar\alpha_{t-1}}\,\hat{x}_0
+\sqrt{1 - \bar\alpha_{t-1} - \sigma_t^2}\;\hat\epsilon
+\sigma_t n
\end{equation}
where $\sigma_t$ is the noise standard deviation at timestep $t$ and $n \sim \mathcal{N}(0,\mathbf{I})$.
To ensure that the same input always yields the same output, we set \(\sigma_t=0\) for all \(t\), thereby making the reverse process deterministic given a fixed conditioning.
}

\subsection{Network Architecture}
\subsubsection{Diffuse Reflection Autoencoder}{
The diffuse reflection autoencoder uses a standard VAE-style encoder--decoder architecture\cite{kingma2019introduction}, re-parameterized for 24-channel phase-shifted fringe stacks.
The input to the autoencoder is a stack of 24 single-channel phase-shifted fringe images captured on a diffuse surface, and the decoder reconstructs a corresponding 24-channel output. Both encoder and decoder use residual blocks with Group Normalization and Swish activations. The encoder begins with a $3\times3$ convolution that maps the input image to a base channel dimension of 64, and then processes the feature map through a cascade of top-level blocks. Each block contains two ResNet blocks and a down-sampling layer except the final block. 
Channel widths grow according to the channel-multiplier vector $\mathbf{m} = (1, 2, 4, 4)$.
A mid-module with two ResNet blocks and an attention block is inserted between encoder and decoder. The encoder output is projected by a $3\times3$ convolution to 8 channels, which parametrize the mean and log-variance of a Gaussian latent distribution.

The decoder mirrors the encoder: an initial $3\times3$ convolution maps latent channels to feature channels, followed by the mid ResNet+attention blocks and top-level up-sampling blocks that restore spatial resolution. 
Two $1\times1$ projection convolutions map between the encoder feature and the latent parameterization and map the latent code back to the decoder input, respectively.
A ResNet block consists of GroupNorm followed by a Swish activation and a $3\times3$ convolution, then another GroupNorm, Swish, and $3\times3$ convolution. Attention is implemented via $1\times1$ convolutions for queries, keys, and values, a spatial softmax, a $1\times1$ projection, and a residual connection. Upsampling is implemented with nearest-neighbor interpolation followed by a $3\times3$ convolution, while downsampling is implemented by zero-padding followed by a stride-2 $3\times3$ convolution. 

During autoencoder pretraining, the encoder predicts the mean and log-variance of a Gaussian latent distribution.
A latent sample can be obtained as
\begin{equation}
z = \mu + \sigma \odot \epsilon, \quad \epsilon \sim \mathcal{N}(0, I).
\end{equation}
For diffusion training and inference, we use the posterior mean as a deterministic latent code, which is then subjected to the forward diffusion process.
}
\subsubsection{Specular Reflection Encoder}{
The direction of specular reflection is determined by the surface normal, so the observed fringe patterns on specular surfaces depend on object geometry and surface reflectance.
To capture these effects, we introduce a Specular Reflection Encoder that extracts characteristic features from fringe images formed by specular reflections. These features are then approximated as a multivariate Gaussian distribution to structure the continuous latent space, enabling both the interpolation between different object properties and a compact representation of the geometric and reflective properties.

The specular encoder shares the architecture of the Diffuse Reflection Autoencoder and is pretrained in the same way: it reconstructs 6-step phase-shifted single channel fringe images captured on specular surfaces and thereby learns an embedding that encodes both shape and reflectance information.
The resulting embedded latent feature map is subsequently used as conditioning input for a denoiser network to aid the denoising process. Since the conditioning information is provided as a persistent input, it is excluded from the forward diffusion process.
Given that the fringe images captured on a specular surface are highly sensitive to object geometry, the spatial information within the latent map significantly contributes to effective conditioning. Therefore, the latent map from the Specular Reflection Encoder is concatenated channel-wise with the latent map from the Diffuse Reflection Autoencoder before being fed into the denoiser network in training.
}
\subsubsection{Denoiser Network}{
We use a U\mbox{-}Net--style denoiser operating in the latent space. 
The input to the denoiser has 8 channels, obtained by concatenating the ground-truth latent and the conditional latent, and the network predicts a 4-channel clean latent $\hat{z}_{d(0)}$. 
For both branches, we use the posterior mean of the encoders as the latent code and do not sample from the predicted variance, ensuring that a fixed conditional input always produces a deterministic target. 
The base channel width is $C=320$. We employ two residual blocks at each resolution level and scale channels according to the multiplier vector $\mathbf{m} = (1, 2, 4, 4)$.
The diffusion timestep $t$ is encoded using a sinusoidal embedding and mapped by a two-layer MLP of width $4C$ to obtain a time embedding $e_t$, which is injected into all residual blocks.

In the input block, we introduce a time-variant channel-wise latent affine modulation, as shown in Fig.~\ref{network}. 
The channel affine module takes the input latent $x_t$ and the time embedding $e_t$ and predicts per-channel scale and shift parameters $\alpha_{b,c}(t)$ and $\beta_{b,c}(t)$, which are broadcast over spatial locations and applied as
\begin{equation}
    \tilde{x}_t(b,c,h,w) = x_t(b,c,h,w)\,\alpha_{b,c}(t) + \beta_{b,c}(t).
\end{equation}
Since the denoiser processes concatenated latents from diffuse and specular encoders and the latent statistics vary with the diffusion timestep, this channel-wise latent affine modulation allows the network to adaptively re-normalize and gate latent channels as a function of both the noise level and the reflection component before entering the main U\mbox{-}Net hierarchy.
The modulated latent $\tilde{x}_t$ is then passed through a $3{\times}3$ convolution to project the channels to the base width $C$, followed by a channel attention block that further reweights the channels.

Beyond the input block, we employ time-independent channel attention modules before each down-sampling in the encoder and after each up-sampling in the decoder. 
These modules apply spatially varying per-channel gating, so that features related to specular distortions and fringe restoration are emphasized consistently across scales.
Each channel attention block reshapes the feature map to a sequence of $HW$ tokens with $C$ channels, applies a small two-layer MLP to the channel dimension followed by a sigmoid activation along the channel dimension, and reshapes the result back to $[B,C,H,W]$ to obtain multiplicative channel gates with smoothly bounded values.

Instead of inserting spatial self-attention at multiple resolutions, we introduce a single spatial self-attention block at the bottleneck to reduce the computational cost.
The middle block consists of a ResNet block, a spatial self attention block, and a second ResNet block. 
The spatial self-attention block normalizes the feature map, applies a $1{\times}1$ projection, flattens the spatial dimensions to a sequence of $HW$ tokens, and processes them with a stack of transformer layers before projecting back to $[B,C,H,W]$ with a residual connection. 
This bottleneck self-attention provides global spatial context in the highest-level latent representation at a modest computational cost.

The remainder of the denoiser follows a standard U\mbox{-}Net--style design with ResNet blocks and skip connections. 
The encoder path consists of ResNet blocks with time-step conditioning, interleaved with strided $3{\times}3$ convolutions for down-sampling. 
The decoder mirrors this structure using nearest-neighbor up-sampling followed by $3{\times}3$ convolutions, and at each resolution it concatenates the corresponding encoder features via skip connections. 
Finally, a GroupNorm--SiLU--$3{\times}3$ convolution head maps the last feature map to a 4-channel clean latent prediction $\hat{z}_{d(0)}$.
The overall training and inference procedures of the denoiser are summarized in Algorithm~\ref{alg:training} and Algorithm~\ref{alg:inference_ddim}, respectively, and both processing flows are also illustrated in Fig.~\ref{network}.
\begin{algorithm}[!t]
\caption{Training}\label{alg:training}
\begin{algorithmic}[1]
\REPEAT
  \STATE $X, Y \sim q_{\mathrm{data}}(X,Y)$
  \STATE $z_{s} = f_{\text{specular-encoder}}(X)$
  \STATE $z_{d(0)} = f_{\text{diffuse-encoder}}(Y)$
  \STATE $t \sim \mathrm{Uniform}(\{1,\ldots,T\})$
  \STATE $\epsilon \sim \mathcal{N}(0, I)$
  \STATE $z_{d(t)} = \sqrt{\bar{\alpha}_t}\,z_{d(0)} + \sqrt{1-\bar{\alpha}_t}\,\epsilon$
  \STATE Take gradient descent step on
  \[
    \nabla
    \big\| x_0 - f_{\text{denoiser}}(z_{d(t)}, z_s, t) \big\|^2 
  \]
\UNTIL converged
\end{algorithmic}
\end{algorithm}

\begin{algorithm}[!t]
\caption{Inference}\label{alg:inference_ddim}
\begin{algorithmic}[1]
\STATE $X \sim q_{\mathrm{data}}(X), \ \tilde{z}_{d(T)} \sim \mathcal{N}(0, I)$
\STATE $z_s = f_{\text{specular-encoder}}(X)$
\FOR{$t = T,\ldots,1$}
  \STATE \textbf{predict clean latent:}\;
  $\hat z_{d(0)} \;=\; f_{\text{denoiser}}(\tilde z_{d(t)}, z_s, t)$
  \vspace{2mm}
  \STATE \textbf{predict noise:}\;
  $\displaystyle \hat\epsilon = \frac{\tilde z_{d(t)} - \sqrt{\bar\alpha_t}\,\hat z_{d(0)}}{\sqrt{1-\bar\alpha_t}} $
  \[
    \tilde z_{d(t-1)}
    \;=\;
    \sqrt{\bar\alpha_{t-1}}\,\hat z_{d(0)}
    \;+\;
    \sqrt{\,1-\bar\alpha_{t-1}-\sigma_t^2\,}\;\hat\epsilon
  \]
\ENDFOR
\STATE $\hat{Y} = f_{\text{diffuse-decoder}}(\tilde z_{d(0)})$
\RETURN $\hat{Y}$
\end{algorithmic}
\end{algorithm}
}

\subsection{Loss Function}{
For pre-training the VAE-based modules\cite{kingma2019introduction}, we employ a composite loss that combines a Huber reconstruction term, a structural similarity (SSIM) term\cite{wang2004imageSSIM}, and a weighted Kullback–Leibler (KL) divergence. The Huber loss is used instead of pure $L_1$ or $L_2$ loss to improve training stability and robustness to outliers in fringe intensity. Since the information in phase-shifted fringe images is largely encoded in local contrast and structural variations, we additionally incorporate an SSIM loss to better preserve fringe quality and fine structural details in the reconstructed images. The overall VAE loss is as follows:
\begin{equation}
\mathcal{L}_{\text{VAE}}
= \mathcal{L}_{\text{Huber}}(x,\hat{x})
+ \mathcal{L}_{\text{SSIM}}(x,\hat{x})
+ \beta\,D_{\mathrm{KL}}\big(q_\phi(z\mid x)\,\|\,p(z)\big),
\label{eq:vae_loss}
\end{equation}
where $q_\phi(z\mid x)$ is the approximate posterior distribution parameterized by the encoder, and $p(z)$ is the standard normal prior.
We adopt different KL weights $\beta$, setting $\beta=10^{-1}$ for the Specular Reflection Encoder
and $\beta=10^{-5}$ for the Diffuse Reflection Autoencoder. The Diffuse Reflection Autoencoder is responsible for generating the final high-quality fringe images used for 3-D reconstruction; thus, we use a relatively small $\beta$ so that the reconstruction terms (Huber and SSIM) dominate the objective, prioritizing perceptual and photometric fidelity over latent regularization. In contrast, the Specular Reflection Encoder serves as a conditional VAE whose primary role is to extract informative latent representations of specular reflection patterns rather than to produce visually accurate reconstructions. For this module, we increase $\beta$ to place more emphasis on the KL divergence, encouraging the latent space to compactly encode and disentangle geometry- and reflectance-related features of specular fringes.
These latent representations can be exploited more effectively as conditioning information by the denoiser network.
}

\subsection{Fringe Pattern Design and Data Preparation}{
\begin{figure}[!t]
\centerline{\includegraphics[width=\columnwidth]{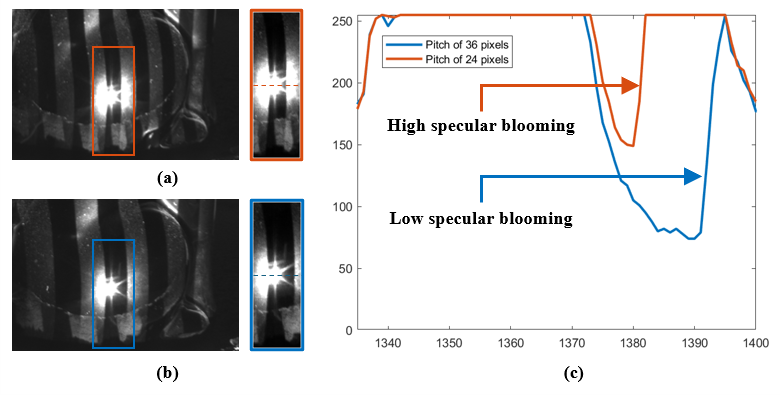}}
\caption{Effect of fringe pitch on specular blooming for highly reflective surfaces. (a) Captured binary fringe image with a pitch of 24 pixels. (b) Captured binary fringe image with a pitch of 36 pixels. (c) Intensity profiles indicating stronger blooming for the 24-pixel pattern and reduced blooming for the 36-pixel pattern.}
\label{fig:specular_blooming}
\end{figure}
We employ two fringe-pattern configurations: one for the network input on highly reflective surfaces and one for generating ground-truth targets. For the input, we project 6-step phase-shifted binary fringes with a pitch of 36 pixels. On plated or mirror-like surfaces, sinusoidal fringe patterns tend to wash out due to specular reflection, whereas binary black-and-white patterns maintain higher effective contrast. The projector focus is carefully adjusted to the object surface so that the binary edges remain sharp, making the fringe transitions clearly observable even under strong specular reflection. In addition, each pattern is recorded at three exposure times; increasing the number of phase steps would linearly increase the acquisition time under multi-exposure capture.
The input sequence is used to condition the encoder rather than to directly compute an accurate phase. Therefore, we select a compact 6-step set to reduce acquisition time while retaining representative fringe transitions.
We choose a coarser pitch of 36 pixels because lower spatial-frequency fringes are typically less affected by specular blooming, leading to fewer overexposed pixels per period than finer-pitch patterns, as shown in Fig. \ref{fig:specular_blooming}.

For ground-truth acquisition, we temporarily apply a removable matte spray to the object surface to obtain a reliable phase reference under diffuse reflection.
To keep the supervision consistent with the projector’s binary operating mode, we also use binary patterns for the ground-truth capture as well. Specifically, we project a 24-step binary fringe sequence with a finer spatial pitch of 24 pixels. The phase is shifted by one pixel per step, which is the minimum discrete shift and provides dense phase sampling. This configuration yields well-formed captured fringes and a high-quality phase map, which we use as the supervisory target for training.

For network input, fringe patterns are captured at three exposure times (0.04\,s, 0.16\,s, 0.64\,s) and fused into high-dynamic-range fringe images, which are then provided to the network. To improve generalization, data augmentation includes horizontal and vertical flips and small rotations ($\pm 3^\circ$).
}

\section{Experiment}
\begin{figure*}[!t]
\centerline{\includegraphics[width=0.85\textwidth]{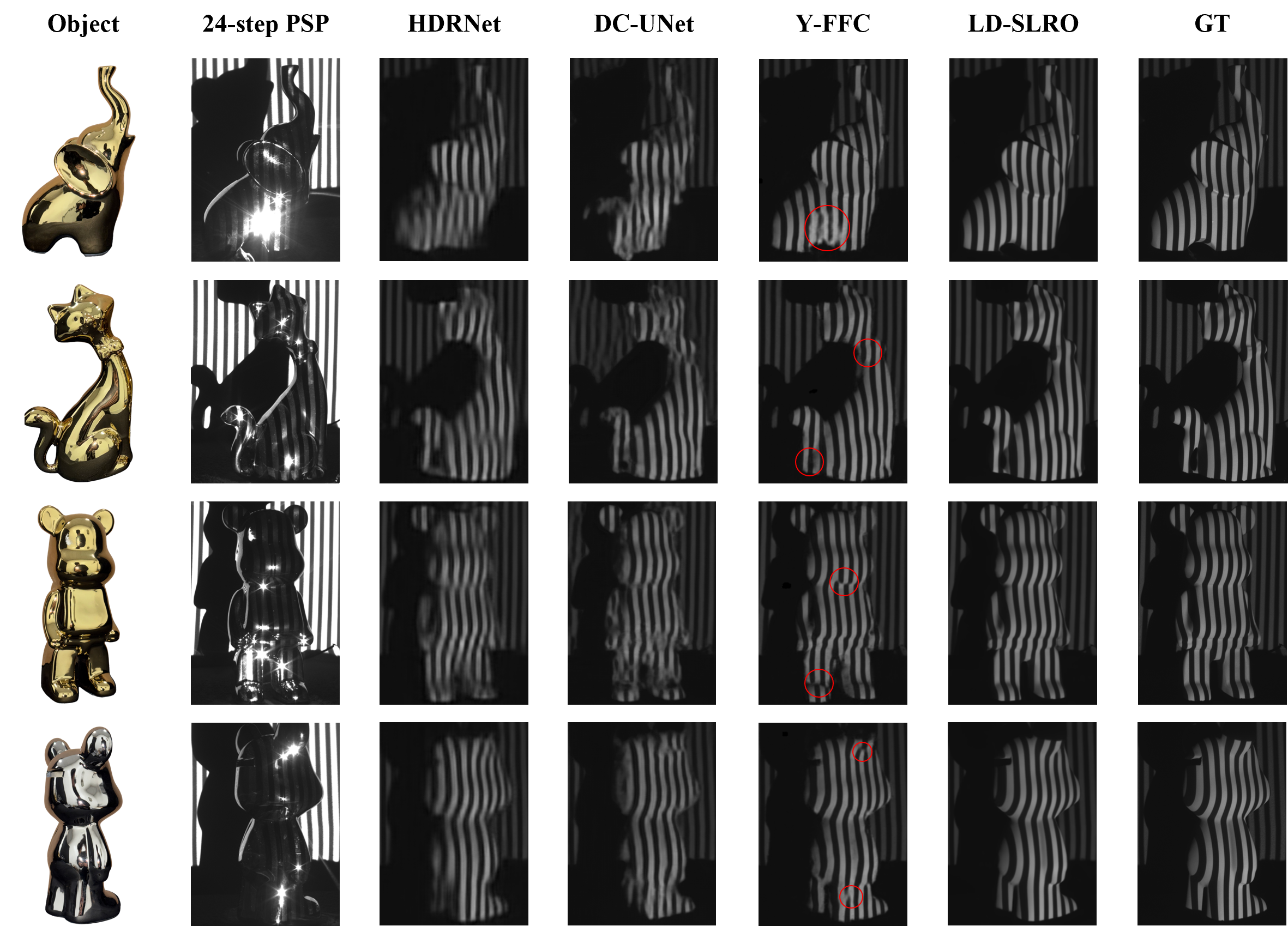}}
\caption{Comparison of fringe image enhancement on highly reflective surfaces using 24-step PSP, HDRNet, DC-UNet, Y-FFC, and the proposed LD-SLRO. The proposed method more effectively suppresses specular highlights and restores high-quality fringe patterns, yielding clearer fringe patterns in overexposure regions.
}
\label{fig:exp1}
\end{figure*}
\subsection{Experimental Setup}
To validate the effectiveness of the proposed method, we built a structured-light measurement system. The system consists of a monochrome CMOS camera (FLIR Grasshopper3 GS3-U3-32S4M) equipped with a 16\,mm focal length lens (Computar M1614-MP2) and a digital light processing (DLP) projector (Texas Instruments LightCrafter 4500). The projector operates at a resolution of $912\times1140$ pixels, while the camera resolution is set to $1920\times1400$ pixels. All experiments were conducted in a darkened environment with blackout curtains to suppress ambient and indirect illumination, so that the captured fringes were dominated by the projected patterns.

Using this setup, we acquired 200 real measurement sets of fringe images, yielding a total of 8,400 fringe images for training and evaluation. The dataset was split into 80\% for training, 10\% for validation, and 10\% for testing. For the ground-truth data, the reflective objects were temporarily coated with a matting spray to reduce specular reflections and approximate a diffuse surface before acquisition. The proposed networks were implemented in PyTorch, and all models were trained on an NVIDIA GeForce RTX 4090 GPU with a learning rate of $2\times10^{-5}$.

\subsection{Ablation Experiment}

\begin{table}[t]
\centering
\caption{Ablation experiment on the proposed components, evaluated using MSE, SSIM, and PSNR on enhanced fringe patterns}
\label{tab:ablation}
\begin{tabular}{cccccc}
\toprule
Encoder & Attention & Affine Layer & MSE $\downarrow$ & SSIM $\uparrow$ & PSNR $\uparrow$ \\
\midrule
\xmark & \xmark & \xmark & 0.0085 & 0.6780 & 20.8948 \\
\cmark & \xmark & \xmark & 0.0004 & 0.9685 & 35.0074 \\
\cmark & \cmark & \xmark & 0.0003 & 0.9698 & 35.9314 \\
\cmark & \cmark & \cmark & \textbf{0.0002} & \textbf{0.9708} & \textbf{36.2924} \\
\bottomrule
\end{tabular}
\end{table}

We conduct ablation experiments to evaluate the contribution of each component of the proposed denoiser network.
The comparison varies three factors in the network architecture: the use of diffuse and specular latent encoders, the use of attention modules, and the presence of the time-variant channel-wise affine layer in the input block. 
All models are trained using the same settings, and fringe enhancement performance is evaluated on the test set using MSE, SSIM, and PSNR. The results are summarized in Table~\ref{tab:ablation}, and the best score for each metric is highlighted in bold.

The baseline model operates directly in image space without any encoder. 
Since it does not learn a compressed latent representation, both training and inference are slow, and this model yields the worst overall. 
Introducing the diffuse and specular latent encoders produces the largest improvement. 
By extracting and compressing reflection-related structures into a low-dimensional latent space, the denoiser becomes more accurate while also reducing the computational cost.

Adding attention modules further improves the reconstruction quality. 
Channel attention and the bottleneck self-attention encourage the network to focus on informative latent features that are critical for correcting fringe distortions and restoring regions affected by overexposure and fringe contrast loss.
Finally, enabling the time-variant channel-wise affine layer in the input block yields an additional improvement. 
This affine modulation helps the network handle the distribution gap between the two encoders and the timestep-dependent change in latent statistics. 
As shown in Table~\ref{tab:ablation}, the full model that includes encoders, attention, and the channel-wise affine layer achieves the best overall performance across all three metrics.

\subsection{Evaluation of Fringe Image Enhancement} {
\begin{table}[t]
\centering
\caption{Quantitative comparison of enhanced fringe images from different methods in terms of MSE, SSIM, and PSNR}
\label{tab:method_comparison}
\small
\setlength{\tabcolsep}{6pt}
\renewcommand{\arraystretch}{1.1}
\begin{tabular}{cccc}
\toprule
Method & MSE $\downarrow$ & SSIM $\uparrow$ & PSNR $\uparrow$ \\
\midrule
24-step PSP       & 0.1747 & 0.5554 & 7.5875 \\
HDRNet       & 0.0015 & 0.9378 & 29.0117 \\
DC-UNet       & 0.0021 & 0.9228 & 27.3877 \\
Y-FFC      & 0.0005 & 0.9634 & 33.3983 \\
LD-SLRO (Ours) & \textbf{0.0002} & \textbf{0.9708} & \textbf{36.2924} \\
\bottomrule
\end{tabular}
\end{table}

\begin{figure*}[!t]
\centerline{\includegraphics[width=0.75\textwidth]{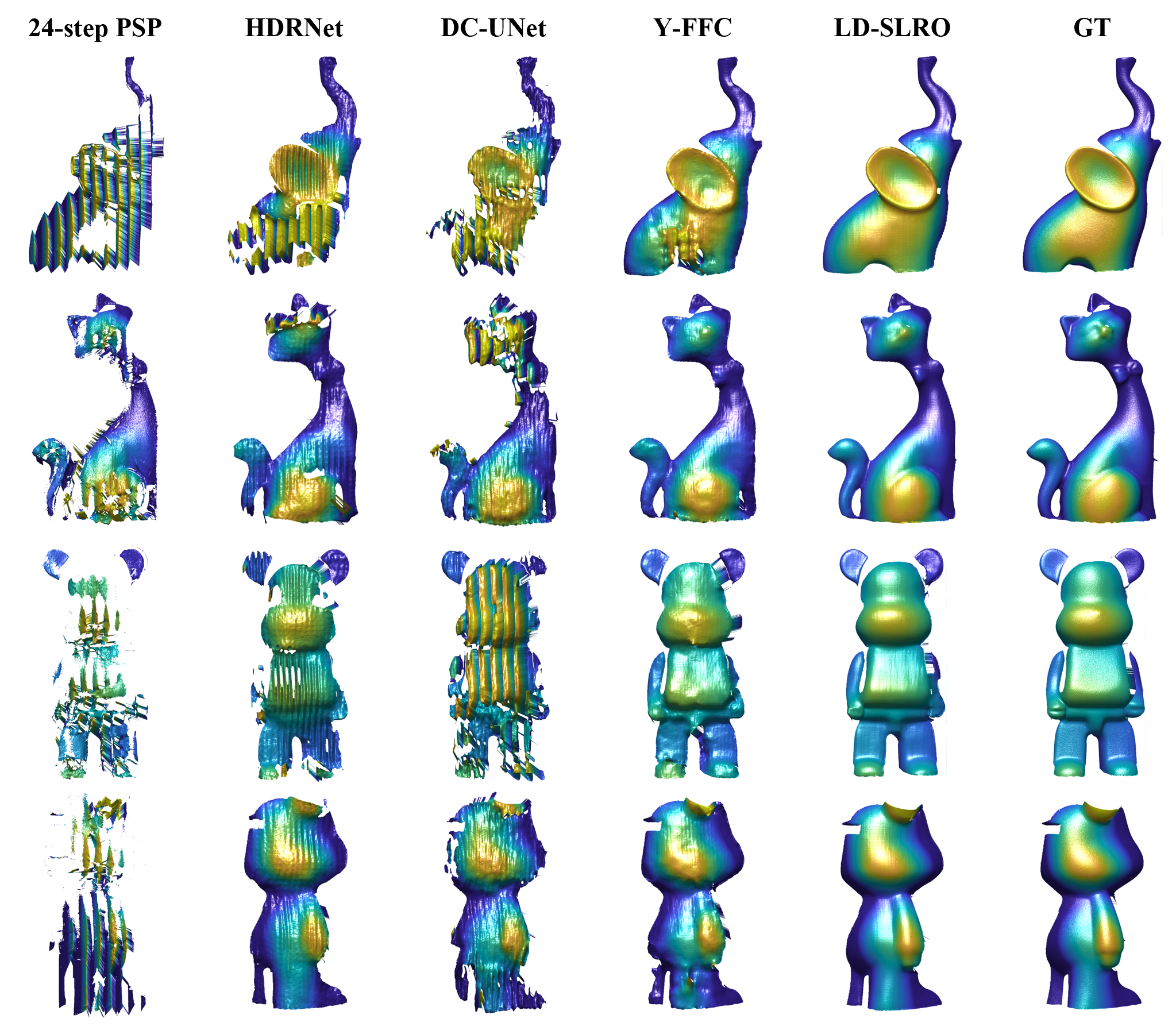}}
\caption{Comparison of 3-D reconstruction results on highly reflective surfaces obtained from the fringe images produced by 24-step PSP, HDRNet, DC-UNet, Y-FFC, and the proposed LD-SLRO. LD-SLRO yields fewer reflection-induced artifacts and more consistent surface geometry in regions affected by overexposure and fringe distortion.
}
\label{fig:exp2}
\end{figure*}

To validate the superiority of the proposed LD-SLRO, we compare it with a conventional method, 24-step phase-shifting profilometry (24-step PSP), and three learning-based methods, including HDRNet\cite{yang2022high-hdrnet}, DC-UNet\cite{li2022three-dcunet}, and Y-FFC\cite{song2024ffc}, which is a recent state-of-the-art network for high-reflection artifact removal. In this section, we evaluate the capability of each method to reduces specular reflection induced errors in the captured fringes, including fringe distortion, fringe contrast loss, and overexposure.

Fig.~\ref{fig:exp1} presents representative enhancement results on highly reflective surfaces. For easier visual inspection, we highlight challenging regions with red circles, where strong specularities and indirect illumination cause severe fringe corruption. As shown in Fig.~\ref{fig:exp1}, HDRNet and DC-UNet tend to over-smooth the fringe structures and fail to remove specular-induced artifacts, leading to blurred or distorted fringe waveforms. Y-FFC improves fringe contrast and reduces high reflection in several regions, but fringe distortions remain in areas dominated by strong reflections. In contrast, the proposed LD-SLRO more effectively suppresses specular induced errors while preserving fine fringe structures, producing results that are the most consistent with the ground truth (GT).

For quantitative evaluation, we compute MSE, SSIM, and peak signal-to-noise ratio (PSNR) between the enhanced fringe images and the ground-truth targets. The comparison is summarized in Table~\ref{tab:method_comparison}, and the best score for each metric is highlighted in bold. Among the learning-based methods, DC-UNet shows the weakest performance. In contrast, the proposed LD-SLRO achieves the best results across all metrics, with an MSE of 0.0002, an SSIM of 0.9708, and a PSNR of 36.2924, indicating its superior capability to restore high-quality fringe patterns on highly reflective surfaces.
}

\subsection{Evaluation of 3-D Reconstruction Accuracy}{
\begin{table}[!t]
\centering
\caption{Comparison of 3-D reconstruction accuracy of different methods in terms of RMSE (mm) for four reflective objects}
\label{tab:rmse_3d}
\small
\setlength{\tabcolsep}{5pt}
\renewcommand{\arraystretch}{1.1}
\begin{tabular}{ccccc}
\toprule
\multirow{2}{*}[-0.3ex]{Method} & \multicolumn{4}{c}{RMSE (mm) $\downarrow$} \\
\cmidrule(lr){2-5}
 & Elephant & Cat & Gold bear & Silver bear \\
\midrule
24-step PSP & 7.4923 & 4.5926 & 5.0371 & 7.6003 \\
HDRNet & 7.3861 & 3.2830 & 3.6585 & 2.0764 \\
DC-UNet & 6.0791 & 6.6130 & 11.8513 & 3.1367 \\
Y-FFC & 1.4929 & 1.1830 & 2.4298 & 2.1646 \\
LD-SLRO & \textbf{0.8059} & \textbf{0.8810} & \textbf{1.3944} & \textbf{0.7666} \\
\bottomrule
\end{tabular}
\end{table}

\begin{figure}[!t]
\centerline{\includegraphics[width=0.8\columnwidth]{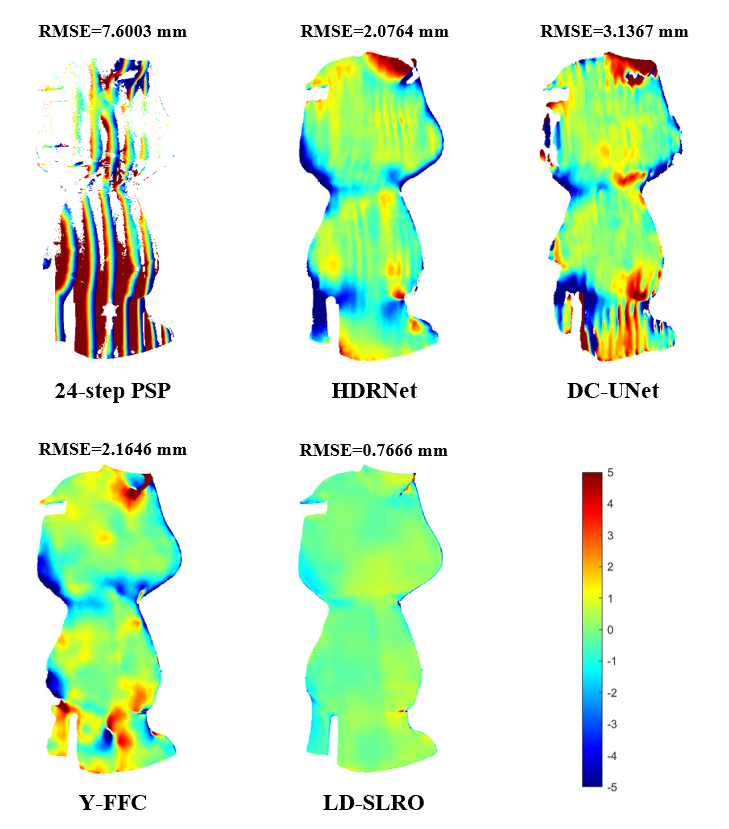}}
\caption{Depth error maps for a representative sample across different methods, shown with a shared color scale.
}
\label{fig:errormap}
\end{figure}
In this section, we compare the 3-D reconstruction accuracy obtained from fringe images produced by five methods: the conventional 24-step phase-shifting profilometry baseline (24-step PSP) and four learning-based methods (HDRNet, DC-UNet, Y-FFC, and the proposed LD-SLRO). For 24-step PSP, the wrapped phase is computed directly from the captured fringe images. In contrast, the learning-based methods first generate enhanced fringe images by reducing reflection-induced error, as shown in Fig.~\ref{fig:exp1}, and the wrapped phase is estimated from these enhanced fringes.
A practical difference among the learning-based methods is the flexibility of the input and output fringe configuration. HDRNet, DC-UNet, and LD-SLRO can take a reduced set of input fringes and synthesize a higher-quality target set with a different number of phase steps and a different fringe pitch, which is useful for accelerating acquisition. In comparison, Y-FFC requires a one-to-one correspondence between the input and output fringes, enforcing the same number of patterns and the same pitch.

The 3-D reconstruction accuracy is sensitive to errors introduced during phase unwrapping. To obtain the absolute phase map and ensure a fair comparison under identical conditions across all methods, we use a geometric-constraint–based phase unwrapping approach\cite{an2016pixel} that requires no additional projection patterns and is thus not affected by specular-reflection–induced errors. The 3-D point cloud is then reconstructed via triangulation.
Fig.~\ref{fig:exp2} presents representative 3-D reconstruction results, and Table~\ref{tab:rmse_3d} reports the root-mean-squared error (RMSE) for quantitative comparison, where the best results are highlighted in bold. The depth error maps with respect to the ground truth are presented in Fig.~\ref{fig:errormap}.
As shown in Fig.~\ref{fig:exp2}, 24-step PSP fails in many regions and produces large errors on highly reflective surfaces. Among the learning-based methods, DC-UNet yields the poorest reconstruction quality, and HDRNet performs slightly better but still shows substantial depth errors. Y-FFC produces results closer to the proposed method, yet noticeable errors remain in regions affected by overexposure and interreflections, and the overall surface quality is lower than that of LD-SLRO.

LD-SLRO achieves the highest reconstruction accuracy across all four objects, with RMSE values of 0.8059\,mm, 0.8810\,mm, 1.3944\,mm, and 0.7666\,mm, outperforming all other methods including Y-FFC. While some high-frequency surface details are not fully preserved in a few challenging regions, the proposed method reduces reflection-induced phase errors and yields the more accurate 3-D reconstructions under severe specular reflections.
}
\section{Conclusion}
This article presents LD-SLRO, a latent diffusion-based fringe restoration framework for structured-light 3-D measurement of highly reflective, low-roughness surfaces. The proposed network restores high-quality fringe images by leveraging a specular reflection encoder and a diffuse reflection encoder to extract reflection-aware latent features, and by guiding the denoiser with attention modules and affine modulation. By improving fringes quality before phase computation, LD-SLRO reduces specular-reflection-induced phase errors and improves reconstruction accuracy in challenging regions affected by overexposure and interreflections. Experimental results verify that the proposed method effectively mitigates reflection-induced fringe corruption and achieves more accurate 3-D reconstruction than competing baselines. LD-SLRO is particularly useful in industrial inspection and automation pipelines with fixed parts or repetitive processes, where a strong diffusion prior can be exploited for stable and reliable measurements under severe reflections.
\bibliography{ref}
\bibliographystyle{IEEEtran}

\end{document}